\documentclass[runningheads]{llncs}
\usepackage[T1]{fontenc}
\usepackage{pifont}
\usepackage{hyperref}
\usepackage{graphicx}
\usepackage{amsmath}
\usepackage{amsfonts}
\usepackage{amssymb}
\usepackage{mhchem}
\usepackage{stmaryrd}
\usepackage{graphicx}
\usepackage[export]{adjustbox}
\usepackage{booktabs, multirow} 
\usepackage{soul}
\usepackage[table]{xcolor} 
\usepackage{changepage,threeparttable} 
\usepackage{multirow}
\newcommand{\repeatthanks}{\textsuperscript{\thefootnote}}

\begin{document}
\title{L3Cube-MahaSBERT and HindSBERT: Sentence BERT Models and Benchmarking BERT Sentence Representations for Hindi and Marathi}
%
\titlerunning{L3Cube-MahaSBERT and HindSBERT: Sentence BERT models}

\author{Ananya Joshi \inst{1,3}\thanks{Authors contributed equally},
Aditi Kajale \inst{1,3}\repeatthanks,
Janhavi Gadre\inst{1,3}\repeatthanks,
Samruddhi Deode\inst{1,3}\repeatthanks\\  and
Raviraj Joshi \inst{2,3}}

\authorrunning{A. Joshi et al.}

\institute{MKSSS' Cummins College of Engineering for Women, Pune, Maharashtra, India \and
Indian Institute of Technology Madras, Chennai, India
\and L3Cube, Pune\\
\vspace{10pt}
\email{\{joshiananya20,aditi1.y.kajale,janhavi.gadre,samruddhi321\}@gmail.com}
\\
\email{ravirajoshi@gmail.com}
}

\maketitle             

\begin{abstract}
Sentence representation from vanilla BERT models does not work well on sentence similarity tasks. Sentence-BERT models specifically trained on STS or NLI datasets are shown to provide 
state-of-the-art performance. However, building these models for low-resource languages is not straightforward due to the lack of these specialized datasets. This work focuses on two low-resource Indian languages, Hindi and Marathi. We train sentence-BERT models for these languages using synthetic NLI and STS datasets prepared using machine translation. We show that the strategy of NLI pre-training followed by STSb fine-tuning is effective in generating high-performance sentence-similarity models for Hindi and Marathi. The vanilla BERT models trained using this simple strategy outperform the multilingual LaBSE trained using a complex training strategy. These models are evaluated on downstream text classification and similarity tasks. We evaluate these models on real text classification datasets to show embeddings obtained from synthetic data training are generalizable to real datasets as well and thus represent an effective training strategy for low-resource languages. We also provide a comparative analysis of sentence embeddings from fast text models, multilingual BERT models (mBERT, IndicBERT, xlm-RoBERTa, MuRIL), multilingual sentence embedding models (LASER, LaBSE), and monolingual BERT models based on L3Cube-MahaBERT and HindBERT. We release L3Cube-MahaSBERT and HindSBERT, the state-of-the-art sentence-BERT models for Marathi and Hindi respectively. Our work also serves as a guide to building low-resource sentence embedding models.


\end{abstract}

\section{Introduction}
On sentence-pair regression tasks like semantic textual similarity, BERT has achieved a new state-of-the-art performance~\cite{dasgupta2018evaluating}. Semantic similarity aims to find sentences having meaning similar to the target sentence and is used in applications like clustering and semantic search \cite{perone2018evaluation}. Initial approaches based on BERT necessitate feeding both sentences into the network, making the task computationally intensive. Such a BERT design renders it unsuitable for unsupervised tasks like clustering as well as semantic similarity searches. In another naive approach, individual sentences are provided as input to BERT, to derive fixed-size sentence embedding. The average of the BERT output layer known as BERT embedding or the output of its [CLS] token is used. However, previous works have shown such sentence embeddings to be unsuitable for semantic similarity tasks. The average embedding is known to work better than the [CLS] token~\cite{wang2020sbert,ma2019universal,reimers2019sentence,conneau2017supervised}. Alternatively, some works have shown [CLS] to work well when the domain of pre-training and target fine-tuning is the same. More recently, computationally efficient Sentence-BERT models were proposed and shown to work well on similarity-based tasks~\cite{reimers2019sentence}. It is a variant of the standard pre-trained BERT that generates sentence embeddings using siamese and triplet networks.
\\\\
Hence, we present Sentence-BERT models for low-resource languages using synthetic NLI and STS datasets prepared using machine translation. Although monolingual or multilingual BERT models have been available for low-resource languages, Sentence-BERT models are still missing due to the non-availability of specialized NLI and STS datasets.
\\\\
A significant amount of research has been done on the English language \cite{conneau2018you,howard2018universal,cer2018universal} but Indian regional languages like Marathi and Hindi lack sufficient language resources \cite{joshi2019deep,velankar2022mono}. Marathi, having its origin in Sanskrit, is rich in morphology \cite{joshi2022l3cubeMahaNLP}. The fast-growing internet presence of these languages suggests the need for research and development.
\\\\
Building Sentence-BERT models for Marathi and Hindi is complex due to the lack of specialized datasets in these languages. Hence, we use synthetic NLI and STS datasets created using translation for training the Sentence-BERT models. The synthetic datasets are desirable in absence of real datasets due to high translation quality which is also highlighted in~\cite{aggarwal2022indicxnli}. 
\\\\
We perform a comparative study of sentence-BERT models along with FastText models, monolingual \cite{joshi2022l3cube,joshil3cube} and multilingual \cite{khanuja2021muril} BERT models of both Hindi and Marathi languages. The models are compared based on their classification accuracy scores and embedding similarity scores. Classification categorizes a set of data into respective classes and is used to evaluate the discriminative capability of sentence embedding. The embedding similarity score captures the capability of embedding to compute the semantic relatedness of sentences. The sentence embeddings are evaluated on real classification datasets to ensure that the sentence BERT models do not overfit the noise in the synthetic corpus. We show that SBERT models trained on translated corpus perform competitively on classification datasets thus indicating the high quality of translations.

\paragraph{}
Our primary observations and contributions are as follows:
\begin{itemize}

\item[$\bullet$]We show that the FastText-based sentence embedding acts as a strong baseline and performs better than or competitively with most of the popular multi-lingual BERT models like IndicBERT, mBERT, and xlm-RoBERTa.\\
\item[$\bullet$]In the vanilla BERT models, monolingual BERT models based on MahaBERT and HindBERT perform best on classification tasks. Whereas the MuRIL performs best on embedding similarity tasks. Although IndicBERT was shown to have state-of-the-art performance on classification tasks, the embedding quality of this model is very poor. Overall the zero-shot capability of LaBSE is the best among all the models.\\
\item[$\bullet$]We study the effect of single-step and two-step training using NLI and STS datasets on monolingual and multilingual base models and provide a comparative analysis of their performance.\\
\item[$\bullet$]We introduce \textbf{HindSBERT}\footnote[1]{\url{https://huggingface.co/l3cube-pune/hindi-sentence-bert-nli}}\footnote[2]{\url{https://huggingface.co/l3cube-pune/hindi-sentence-similarity-sbert}} and \textbf{MahaSBERT}\footnote[3]{\url{https://huggingface.co/l3cube-pune/marathi-sentence-bert-nli}}\footnote[4]{\url{https://huggingface.co/l3cube-pune/marathi-sentence-similarity-sbert}}, the sentence-BERT models trained using translated versions of the NLI and STSb datasets, which outperform the other models tested in this paper. These models are fine-tuned versions of HindBERT and MahaBERT respectively. To the best of our knowledge, this is the first documented work on dedicated sentence BERT models for low-resource Marathi and Hindi languages.

\end{itemize}
The next part of the paper is structured as follows. Section 2 explores the research work that compares BERT-based models, and analyzes and suggests ways to improve their performance. Previous work related to the development of sentence-BERT models is also surveyed. Section 3 puts forth the details of the datasets used in this work. Section 4.1 describes the models used, section 4.2 describes the experiment and evaluation setup, and section 4.3 explains the findings from our experiments. We conclude the paper with a summary of all the observations in section 5. This work is released as a part of the MahaNLP project \cite{joshi2022l3cubeMahaNLP}.  

\section{Related Work}
BERT~\cite{devlin2018bert} is a pre-trained transformer network, one of the most effective language models in terms of performance when different NLP tasks like text classification are concerned. However, there are some shortcomings in BERT, which have been identified. ~\cite{li2020sentence} investigates the deficiency of the BERT sentence embeddings on semantic textual similarity, and proposes a flow-based calibration which can effectively improve the performance. In~\cite{oh2022don} , the authors introduce the attention-based pooling strategy that enables in preserving of layer-wise signals captured in each layer and learning digested linguistic features for downstream tasks. It demonstrates that training a layer-wise attention layer with contrastive learning objectives outperforms BERT and pre-trained language model variants as well.
\\\\
The previous research~\cite{scheible2020gottbert} has shown how a German monolingual BERT model based on RoBERTa outperforms all other tested German and multilingual BERT models with a little tuning of hyperparameters. Similarly, a Czech monolingual RoBERTa language model has been presented in~\cite{straka2021robeczech}, wherein authors show that the model significantly outperforms equally-sized multilingual and Czech language-oriented model variants. A similar study has been undertaken for the Marathi language as well. ~\cite{velankar2022mono} compares the standard multilingual BERT-based models with their corresponding Marathi monolingual versions. It highlights the superior performance and sentence representations from the monolingual variants over the multilingual ones when focused on single-language tasks. ~\cite{joshi2022l3cube} presents the MahaFT- Marathi fast text embeddings trained on the Marathi monolingual corpus and shows that it performs competitively with other publicly available fast text embeddings. The sentence embeddings of BERT and LASER for the Hindi language have been evaluated in \cite{joshi2019deep}. They report sub-optimal zero-shot performance of these sentence embeddings on a variety of Hindi text classification tasks.
\\\\
In~\cite{feng2020language}, the authors present LaBSE, a language-independent sentence embedding model that supports 109 languages. In comparison to the previous state-of-the-art, the model achieves superior performance on a variety of text retrieval or mining tasks, as well as increased language coverage.  In this work, we perform an extensive evaluation of this model.
\\\\
~\cite{reimers2019sentence} presents the Sentence-BERT (SBERT), which is a computationally efficient and fine-tuned BERT in a siamese or triplet network architecture. The authors show that training on NLI, followed by training on STSb leads to a marginal improvement in performance. Our work is centred around SBERT architecture presented in this work. ~\cite{choi2021evaluation} explores sentence-ALBERT (SALBERT) along with experimenting with CNN sentence-embedding network for SBERT and SALBERT. The Findings of the experiment show that CNN architecture improves AlBERT models for STS benchmark.

\section{Datasets}
This section lists the public datasets utilized in our experiment:
\\\\
\textbf{IndicXNLI\footnote[5]{\url{https://github.com/divyanshuaggarwal/IndicXNLI}}} consists of English XNLI data translated into eleven Indic languages, including Hindi and Marathi~\cite{aggarwal2022indicxnli}. The train (392702), validation (2490), and evaluation sets (5010) of English XNLI are translated from English into each of the eleven Indic languages. From IndicXNLI, we use the training samples of the corresponding language to train the HindSBERT and MahaSBERT.
\\\\
The \textbf{STS benchmark (STSb)\footnote[6]{\url{https://huggingface.co/datasets/stsb\_multi\_mt}}} is a widely used dataset for assessing supervised Semantic Textual Similarity (STS) systems. In STS, sentence pairs are annotated with a score indicating their similarity, with scores ranging from 0 to 5. The data includes 8628 sentence pairs from the three groups- captions, news, and forums. It is divided into 5749 train, 1500 dev and 1379 test~\cite{sun2022sentence}. We translate the STSb dataset using Google Translate to Marathi and Hindi for training and evaluating the MahaSBERT-STS and HindSBERT-STS. It is made accessible publicly\footnote[7]{\url{https://github.com/l3cube-pune/MarathiNLP}}.
\\\\
The down-stream evaluation of BERT-based models is performed on the following Marathi and Hindi classification datasets:
\begin{itemize}
    \item[$\bullet$] \textbf{L3Cube-MahaSent}: A Sentiment Analysis dataset in Marathi that includes tweets classified as positive, negative, and neutral ~\cite{kulkarni2021l3cubemahasent}. The number of train, test, and validation examples are 12114, 2250, and 1500 respectively.\\
    \item[$\bullet$] \textbf{IndicNLP News Articles}: A dataset containing Marathi news articles classified into three categories: sports, entertainment, and lifestyle. The dataset has 4779 records total, of which 3823, 479, and 477 are found in the train, test, and validation sets respectively~\cite{kunchukuttan2020ai4bharat}.\\
    \item[$\bullet$] \textbf{iNLTK Headlines}: A dataset containing Marathi news article headlines from three different categories: entertainment, sports, and state. It has 12092 records, which are divided into 9672 train, 1210 test, and 1210 validation samples.\\
    \item[$\bullet$] \textbf{BBC News Articles}: A corpus of Hindi text classification extracted from the BBC news website. It consists of 3466 train and 865 test samples. 500 samples chosen randomly from the train data are used for validation.\\
    \item[$\bullet$] \textbf{IITP Product reviews}: A sentiment analysis set of Hindi product reviews from 3 classes- positive, negative, and neutral. It contains 4181 training samples, 522 validation, and 522 test samples.\\
    \item[$\bullet$] \textbf{IITP Movie reviews}: A sentiment analysis set of Hindi movie reviews divided into 3 classes- positive, negative, and neutral. It contains 2479 training samples, 309 validation, and 309 test samples.
\end{itemize}

\begin{table}[!htp]\centering
\scalebox{0.8}{
\begin{tabular}{lccccc}\toprule
&\textbf{Dataset} &\textbf{Training  } &\textbf{Validation   } &\textbf{Test} \\\midrule
\multirow{2}{*}{\textbf{Multilingual  }} &IndicXNLI &392702 &2490 &5010 \\
&STSb &5749 &1500 &1379 \\\midrule
\multirow{3}{*}{\textbf{Monolingual-Marathi  }} &L3Cube-MahaSent &12114 &1500 &2250 \\
&Articles &3814 &476 &477 \\
&Headlines &9671 &476 &1209 \\\midrule
\multirow{3}{*}{\textbf{Monolingual-Hindi  }} &BBC News Articles &3466 &500 &865 \\
&IITP- Product reviews &4181 &522 &522 \\
&IITP- Movie reviews &2479 &309 &309 \\
\bottomrule\\
\end{tabular}
}
\caption{Number of samples present in various Hindi and Marathi datasets}\label{tab:1 }
\end{table}
\newpage
\section{Experiments}

\subsection{Models}

\subsubsection{A. FastText models\\}

For morphologically rich languages, FastText word embeddings are commonly used. This method extends the word2vec model by representing the word as a bag of character n-grams, preventing words that are out of vocabulary~\cite{joshi2022l3cube}. The \textbf{L3Cube-MahaFT~\cite{joshi2022l3cube}} is a FastText model trained on a 24.8M sentence and 289M token Marathi monolingual corpus. The \textbf{FB-FT}\footnote[8]{\url{https://fasttext.cc/docs/en/crawl-vectors.html}} is a set of fast text embedding models made available by Facebook by training on Wiki and the Common Crawl Corpus~\cite{grave2018learning}. The FB-FT is available in both Hindi and Marathi languages.

\subsubsection{B. BERT models\\}

The BERT is a deep Bi-directional Transformer-based model trained on a large unlabeled corpus~\cite{devlin2018bert}. A variety of transformer-based pre-trained BERT models is publicly available. We have explored multiple monolingual and multilingual models in this work. We tried three different pooling strategies for each of these models: CLS embeddings, MEAN embeddings, and MAX embeddings from all tokens.
\\\\
Following are the standard multilingual BERT models which use Hindi and Marathi as training languages:
\begin{itemize}
  \item[$\bullet$] \textbf{IndicBERT\footnote[9]{\url{https://huggingface.co/ai4bharat/indic-bert}}}: a multilingual ALBERT model developed by Ai4Bharat trained on a large volume of data. The training languages include 12 major Indian languages~\cite{kakwani2020indicnlpsuite}.\\

  \item[$\bullet$] \textbf{xlm-RoBERTa\footnote[10]{\url{https://huggingface.co/xlm-roberta-base}}}: the RoBERTa model supporting numerous languages. It is pre-trained with the Masked language modelling (MLM) objective on 2.5TB of filtered CommonCrawl data containing 100 languages~\cite{conneau2019unsupervised}.\\

  \item[$\bullet$] \textbf{mBERT\footnote[11]{\url{https://huggingface.co/bert-base-multilingual-cased}}}: a BERT-base model pre-trained on 104 languages using next sentence prediction (NSP) objective and Masked Language Modeling (MLM)~\cite{DBLP:journals/corr/abs-1810-04805}.\\

  \item[$\bullet$] \textbf{MuRIL\footnote[12]{\url{https://huggingface.co/google/muril-base-cased}}} (Multilingual Representations for Indian Languages): a BERT-based model pre-trained on 17 Indic languages and parallel data~\cite{khanuja2021muril} which includes the translations as well as transliterations on each of the 17 monolingual corpora.\\
  
  \item[$\bullet$] \textbf{LaBSE\footnote[13]{\url{https://huggingface.co/setu4993/LaBSE}}} (Language-agnostic BERT sentence embedding): The model~\cite{feng2020language} provides good results while looking for sentence translations. It is trained to output vectors close to each other for pairs of bilingual sentences which are translations of each other.\\
  
  \item[$\bullet$]
  \textbf{LASER\footnote[14]{\url{https://github.com/facebookresearch/LASER}}} (Language-Agnostic Sentence Representations): released by Facebook~\cite{heffernan2022bitext}, provides multilingual sentence representations supporting 90+ languages including low-resource languages like Hindi and Marathi. It uses a single model to handle a variety of languages. This model embeds all languages jointly in a single shared space.

  \end{itemize}

 The following monolingual models are used for comparison with the multilingual models:
  
 \subsubsection{Marathi:}

\begin{itemize}
  \item[$\bullet$] \textbf{MahaBERT\footnote[15]{\url{https://huggingface.co/l3cube-pune/marathi-bert-v2}}}: a multilingual BERT model~\cite{joshi2022l3cube}, fine-tuned using 752M tokens from the L3Cube-MahaCorpus and other freely accessible Marathi monolingual datasets.\\

  \item[$\bullet$] \textbf{MahaAlBERT\footnote[16]{\url{https://huggingface.co/l3cube-pune/marathi-albert-v2}}}: a Marathi monolingual model~\cite{joshi2022l3cube} extended from AlBERT, trained on L3Cube-MahaCorpus and other public Marathi monolingual datasets.\\

  \item[$\bullet$] \textbf{MahaRoBERTa\footnote[17]{\url{https://huggingface.co/l3cube-pune/marathi-roberta}}}: a Marathi RoBERTa model~\cite{joshi2022l3cube} built upon a multilingual RoBERTa model and fine-tuned on publicly available Marathi monolingual datasets including L3Cube-MahaCorpus.\\

  \item[$\bullet$] \textbf{MahaTweetBERT\footnote[18]{\url{https://huggingface.co/l3cube-pune/marathi-tweets-bert}}}: A MahaBERT model~\cite{patankar2022spread} finetuned on the Marathi Tweets dataset.

 \end{itemize}
 
 \subsubsection{Hindi:}

\begin{itemize}
  \item[$\bullet$] \textbf{HindBERT\footnote[19]{\url{https://huggingface.co/l3cube-pune/hindi-bert-v2}}}: It is a multilingual BERT model fine-tuned on publicly available Hindi monolingual datasets~\cite{joshil3cube}.\\

  \item[$\bullet$] \textbf{HindALBERT\footnote[20]{\url{https://huggingface.co/l3cube-pune/hindi-albert}}}: HindAlBERT is a Hindi AlBERT model~\cite{joshil3cube} trained on publicly available Hindi monolingual datasets.\\

  \item[$\bullet$] \textbf{HindRoBERTa\footnote[21]{\url{https://huggingface.co/l3cube-pune/hindi-roberta}}}: It is a multilingual RoBERTa model~\cite{joshil3cube} fine-tuned on publicly available Hindi monolingual datasets.\\

  \item[$\bullet$] \textbf{HindTweetBERT\footnote[22]{\url{https://huggingface.co/l3cube-pune/hindi-tweets-bert}}}: The HindBERT model~\cite{joshil3cube} is finetuned on Hindi Tweets.\\
  
  \item[$\bullet$] \textbf{Sentence-similarity-hindi\footnote[23]{\url{https://huggingface.co/hiiamsid/sentence\_similarity\_hindi}}} This is a sentence-transformer model. It can be used for tasks like clustering or semantic search because it maps sentences and paragraphs to a 768-dimensional dense vector space~\cite{url1}.
  
 \end{itemize}

\subsubsection{C. SBERT models\\} 
\begin{figure}[]
\centering
\includegraphics[scale=0.6]{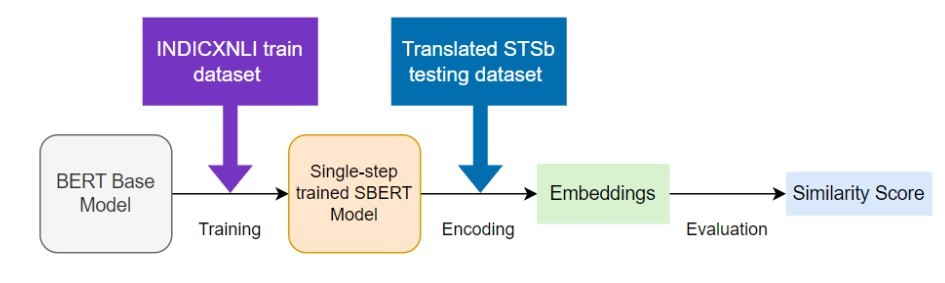}
\caption{Training the BERT base model on the IndicXNLI dataset alone}\label{fig:2}
\end{figure}

\begin{figure}[]
\centering
\includegraphics[scale=0.6]{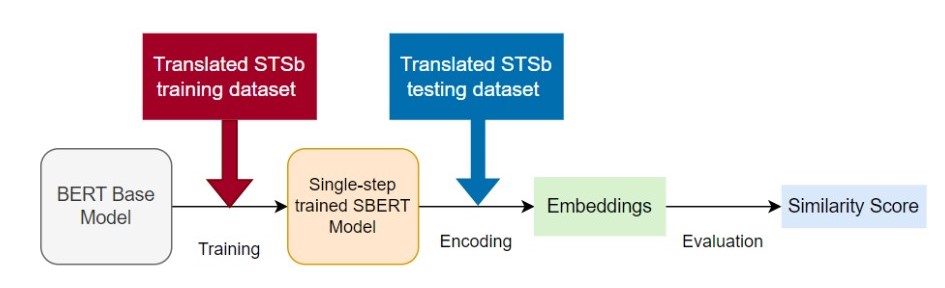}
\caption{Training the BERT base model on the STSb dataset alone}\label{fig:3}
\end{figure}

\begin{figure}[]
\centering
\includegraphics[scale=0.6]{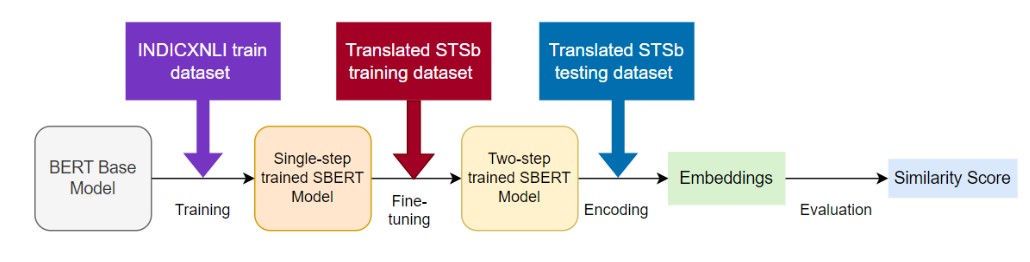}
\caption{Fine-tuning the model trained on IndicXNLI with the translated STSb dataset }\label{fig:4}
\end{figure}

The Sentence-BERT models are created using translated versions of the STSb and NLI datasets. We experiment with two multilingual base models- LaBSE and MuRIL, and the monolingual base model HindBERT for Hindi, and MahaBERT for Marathi.  
\\\\
We experiment on 3 different setups while evaluating the CLS and mean pooling strategies for each setup:
\begin{enumerate}
\item \textbf{Single step} training of the base model on the \textbf{IndicXNLI dataset} alone [fig\ref{fig:2}].  In this method, about 256,180 training sentence triplets (anchor, entailment, contradiction) are used for training with the MultipleNegativesRankingLoss function. Training is done for 1 epoch with batch size 4, AdamW optimizer and a learning rate of 2e-05. The Hindi and Marathi models trained through using setup are termed as HindSBERT and MahaSBERT respectively.\\
\item \textbf{Single step} training of the base model on the translated \textbf{STSb dataset} alone, where combinations of sentence pairs and corresponding similarity scores are used [fig\ref{fig:3}]. Training is done for 4 epochs with CosineSimilarityLoss, which trains the network with a siamese network structure. The AdamW optimizer is used with a learning rate of 2e-05. The base models used for training these models are the HindBERT and MahaBERT respectively.

\begin{figure}[h]
\centering
\includegraphics[scale=0.4]{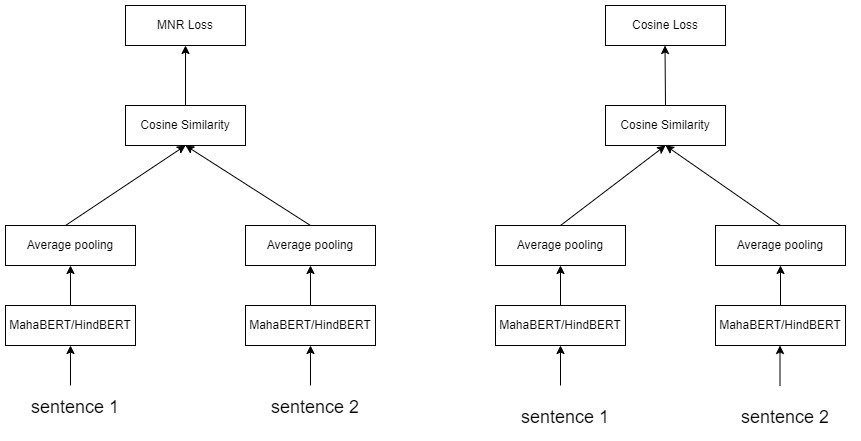}
\caption{Network Structure of single-step trained sentence-BERT}
\end{figure}


\item \textbf{Two step} training, where the SBERT models obtained from setup 1 are fine-tuned using the translated STSb dataset, in 4 epochs and a batch size of 8 [fig \ref{fig:4}]. The AdamW optimizer is used with a learning rate of 2e-05, and the loss function used is CosineSimilarityLoss. The Hindi and Marathi models trained through using setup are termed HindSBERT-STS and MahaSBERT-STS respectively. The base models used for training these models are the HindBERT and MahaBERT respectively.
\end{enumerate}


\subsection{Evaluation Methodology}
We evaluate different monolingual and multilingual BERT models and the sentence-BERT models on their embedding similarity scores and classification accuracies. The embedding Similarity score denotes the Spearman’s Rank Correlation between the cosine similarity scores of the model embeddings and ground truth labels provided in the STSb ~\cite{sun2022sentence}. A high embedding similarity score points to high-quality embeddings in comparison to the benchmark embeddings. 
\\\\
For the text classification datasets, the sentence embeddings are calculated by passing the text through the respective BERT model or FastText model. The embedding-label pair are then classified using the K Nearest Neighbours (KNN) algorithm. It is a non-parametric, supervised learning classifier that leverages proximity to classify or predict how a particular data point will be grouped. The distance metric used is the Minkowski - the generalized form of Euclidean and Manhattan distance metrics. A validation dataset is used to compute the optimal value of k. This value of k is further used to find the accuracy of the test dataset which is reported in this work.
\\\\
We use 3 methods for training the sentence-BERT models, as described in the previous section. In all the 3 methods, the models are tested on translated STSb test dataset to examine their accuracies. We compare these SBERT models with mono and multilingual BERT models on the quality of embeddings generated.


\begin{table}[!htp]\centering
\scalebox{0.8}{
\begin{tabular}{ccccccc}\toprule
\textbf{Model} &\textbf{Pooling} &\textbf{Embedding Similarity} &\textbf{L3Cube-MahaSent} &\textbf{Articles } &\textbf{Headlines} \\\midrule
\multicolumn{6}{c}{\textbf{Fasttext models}} \\\midrule
\textbf{Facebook FB-FT } &\textbf{AVG} &0.53 &0.75 &0.99 &0.9 \\
\textbf{L3Cube MahaFT } &\textbf{AVG} &0.5 &0.76 &0.99 &0.92 \\\midrule
\multicolumn{6}{c}{\textbf{Multilingual variants}} \\\midrule
\multirow{3}{*}{\textbf{IndicBERT}} &\textbf{CLS} &0.12 &0.72 &0.84 &0.73 \\
&\textbf{AVG} &0.35 &\textbf{0.76} &0.9 &\textbf{0.85} \\
&\textbf{MAX} &\textbf{0.36} &0.74 &\textbf{0.92} &0.8 \\\midrule
\multirow{3}{*}{\textbf{xlm-RoBERTa}} &\textbf{CLS} &0.3 &0.72 &\textbf{0.99} &\textbf{0.82} \\
&\textbf{AVG} &\textbf{0.39} &\textbf{0.74} &0.97 &0.81 \\
&\textbf{MAX} &0.35 &0.66 &0.92 &0.72 \\\midrule
\multirow{3}{*}{\textbf{mBERT}} &\textbf{CLS} &0.16 &0.68 &0.94 &0.72 \\
&\textbf{AVG} &0.48 &\textbf{0.7} &\textbf{0.98} &\textbf{0.83} \\
&\textbf{MAX} &\textbf{0.49} &0.67 &0.95 &0.78 \\\midrule
\multirow{3}{*}{\textbf{MuRIL}} &\textbf{CLS} &0.3 &0.72 &\textbf{0.99} &0.9 \\
&\textbf{AVG} &\textbf{0.59} &\textbf{0.78} &0.98 &\textbf{0.91} \\
&\textbf{MAX} &0.5 &0.74 &0.92 &0.8 \\\midrule
\textbf{LASER} & &\textbf{0.62} &0.67 &0.93 &0.73 \\\midrule
\multirow{3}{*}{\textbf{LaBSE}} &\textbf{CLS} &0.7 &\textbf{0.75} &0.98 &0.84 \\
&\textbf{AVG} &\textbf{0.72} &0.75 &0.98 &0.84 \\
&\textbf{MAX} &0.71 &0.73 &\textbf{0.99} &\textbf{0.89} \\\midrule
\multicolumn{6}{c}{\textbf{Monolingual variants}} \\\midrule
\multirow{3}{*}{\textbf{MahaAlBERT}} &\textbf{CLS} &0.27 &0.68 &0.83 &0.81 \\
&\textbf{AVG} &\textbf{0.49} &\textbf{0.77} &0.92 &\textbf{0.86} \\
&\textbf{MAX} &0.48 &0.73 &\textbf{0.93} &0.82 \\\midrule
\multirow{3}{*}{\textbf{MahaTweetBERT}} &\textbf{CLS} &0.26 &0.72 &\textbf{0.99} &\textbf{0.92} \\
&\textbf{AVG} &\textbf{0.53} &\textbf{0.79} &\textbf{0.99} &0.9 \\
&\textbf{MAX} &0.5 &0.76 &0.95 &0.77 \\\midrule
\multirow{3}{*}{\textbf{MahaRoBERTa}} &\textbf{CLS} &0.29 &0.66 &0.98 &\textbf{0.9} \\
&\textbf{AVG} &\textbf{0.55} &\textbf{0.78} &\textbf{0.99} &0.88 \\
&\textbf{MAX} &0.51 &0.72 &0.96 &0.83 \\\midrule
\multirow{3}{*}{\textbf{MahaBERT}} &\textbf{CLS} &0.27 &0.74 &\textbf{0.99} &0.9 \\
&\textbf{AVG} &\textbf{0.55} &\textbf{0.78} &0.98 &\textbf{0.91} \\
&\textbf{MAX} &0.52 &0.76 &0.95 &0.84 \\
\bottomrule\\
\end{tabular}}
\caption{Results for monolingual and multilingual Marathi models}\label{tab:2 }
\end{table}

\begin{table}[!htp]
\centering
\scalebox{0.75}{
\begin{tabular}{ccccccc}\toprule
\textbf{Model} &\textbf{Pooling} &\textbf{Embedding Similarity} &\textbf{News Articles} &\textbf{IITP- Products} &\textbf{IITP- Movies } \\\midrule
\multicolumn{6}{c}{\textbf{Fasttext model}} \\\midrule
\textbf{Facebook FB-FT} &\textbf{AVG} &0.45 &0.67 &0.62 &0.45 \\\midrule
\multicolumn{6}{c}{\textbf{Multilingual variants}} \\\midrule
\multirow{3}{*}{\textbf{IndicBERT}} &\textbf{CLS} &0.15 &0.43 &0.59 &0.47 \\
&\textbf{AVG} &0.34 &\textbf{0.48} &\textbf{0.63} &0.52 \\
&\textbf{MAX} &\textbf{0.37} &0.47 &0.61 &\textbf{0.53} \\\midrule
\multirow{3}{*}{\textbf{mBERT}} &\textbf{CLS} &0.16 &0.55 &0.63 &0.47 \\
&\textbf{AVG} &\textbf{0.48} &\textbf{0.61} &\textbf{0.65} &0.46 \\
&\textbf{MAX} &0.48 &0.5 &0.62 &\textbf{0.5} \\\midrule
\multirow{3}{*}{\textbf{xlm-RoBERTa}} &\textbf{CLS} &0.34 &\textbf{0.64} &\textbf{0.64} &0.48 \\
&\textbf{AVG} &\textbf{0.46} &0.61 &0.64 &\textbf{0.48} \\
&\textbf{MAX} &0.44 &0.49 &0.56 &0.45 \\\midrule
\multirow{3}{*}{\textbf{MuRIL}} &\textbf{CLS} &0.29 &\textbf{0.67} &0.6 &0.45 \\
&\textbf{AVG} &\textbf{0.54} &0.67 &\textbf{0.67} &\textbf{0.55} \\
&\textbf{MAX} &0.47 &0.45 &0.65 &0.51 \\\midrule
\textbf{LASER} & &\textbf{0.65} &0.54 &0.62 &0.5 \\\midrule
\multirow{3}{*}{\textbf{LaBSE}} &\textbf{CLS} &0.7 &0.64 &0.67 &\textbf{0.5} \\
&\textbf{AVG} &\textbf{0.72} &0.66 &\textbf{0.68} &0.48 \\
&\textbf{MAX} &0.71 &\textbf{0.67} &0.67 &0.5 \\\midrule
\multicolumn{6}{c}{\textbf{Monolingual variants}} \\\midrule
\multirow{3}{*}{\textbf{HindAlBERT}} &\textbf{CLS} &0.2 &0.46 &0.6 &0.48 \\
&\textbf{AVG} &0.44 &\textbf{0.56} &0.65 &0.5 \\
&\textbf{MAX} &\textbf{0.47} &0.53 &\textbf{0.66} &\textbf{0.52} \\\midrule
\multirow{3}{*}{\textbf{HindBERT}} &\textbf{CLS} &0.25 &0.67 &0.61 &0.46 \\
&\textbf{AVG} &\textbf{0.52} &\textbf{0.7} &\textbf{0.69} &\textbf{0.53} \\
&\textbf{MAX} &0.5 &0.59 &0.68 &0.51 \\\midrule
\multirow{3}{*}{\textbf{HindTweetBERT}} &\textbf{CLS} &0.18 &0.48 &0.64 &0.49 \\
&\textbf{AVG} &\textbf{0.53} &\textbf{0.66} &\textbf{0.7} &\textbf{0.54} \\
&\textbf{MAX} &0.51 &0.53 &0.61 &0.5 \\\midrule
\multirow{3}{*}{\textbf{HindRoBERTa}} &\textbf{CLS} &0.22 &0.59 &0.6 &0.49 \\
&\textbf{AVG} &0.53 &\textbf{0.69} &\textbf{0.66} &\textbf{0.55} \\
&\textbf{MAX} &\textbf{0.54} &0.59 &0.64 &0.47 \\\midrule
\multirow{3}{*}{\textbf{sentence-similarity-hindi}} &\textbf{CLS} &\textbf{0.82} &0.62 &\textbf{0.75} &0.54 \\
&\textbf{AVG} &\textbf{0.82} &0.63 &0.72 &0.52 \\
&\textbf{MAX} &0.8 &\textbf{0.65} &0.67 &\textbf{0.58} \\
\bottomrule\\
\end{tabular}
}
\caption{Results for monolingual and multilingual Hindi models }\label{tab:4}
\end{table}

\begin{table}[!htp]\centering
\scalebox{0.75}{
\begin{tabular}{cccccc}
\multicolumn{6}{c}{\textbf{Marathi SBERT}} \\\midrule
\textbf{Training setup} &\textbf{Base model} &\textbf{Embedding Similarity} &\textbf{L3Cube-MahaSent} &\textbf{Articles } &\textbf{Headlines} \\\midrule
\multirow{3}{*}{One-step (NLI)} &MuRIL &0.74 &0.8 &0.98 &0.85 \\
&LaBSE &0.76 &0.79 &0.99 &0.82 \\
&MahaBERT &\textbf{0.77} &0.8 &0.98 &0.88 \\\midrule
\multirow{3}{*}{One-step (STS)} &MuRIL &0.77 &0.74 &0.99 &0.89 \\
&LaBSE &\textbf{0.83} &0.78 &0.99 &0.82 \\
&MahaBERT &0.8 &0.79 &0.98 &0.92 \\\midrule
\multirow{3}{*}{Two-step (NLI+STS)} &MuRIL &0.81 &0.79 &0.98 &0.88 \\
&LaBSE &\textbf{0.83} &0.78 &0.99 &0.89 \\
&MahaBERT &\textbf{0.83} &0.79 &0.99 &0.9 \\
& & & & & \\
\multicolumn{6}{c}{\textbf{Hindi SBERT}} \\\midrule
\textbf{Training setup} &\textbf{Base model} &\textbf{Embedding Similarity} &\textbf{News Articles} &\textbf{IITP- Products } &\textbf{IITP- Movies} \\\midrule
\multirow{3}{*}{One-step (NLI)} &MuRIL &0.74 &0.7 &0.7 &0.7 \\
&LaBSE &\textbf{0.75} &0.64 &0.75 &0.56 \\
&HindBERT &0.77 &0.69 &0.75 &0.53 \\\midrule
\multirow{3}{*}{One-step (STS)} &MuRIL &0.79 &0.65 &0.65 &0.65 \\
&LaBSE &\textbf{0.83} &0.65 &0.73 &0.55 \\
&HindBERT &0.82 &0.68 &0.68 &0.48 \\\midrule
\multirow{3}{*}{Two-step (NLI+STS)} &MuRIL &0.83 &0.69 &0.72 &0.49 \\
&LaBSE &0.84 &0.65 &0.74 &0.56 \\
&HindBERT &\textbf{0.85} &0.68 &0.74 &0.5 \\
\bottomrule\\
\end{tabular}
}
\caption{Results of SBERT models}\label{tab:3 }
\end{table}

\begin{table}
\centering
\includegraphics[width=\columnwidth]{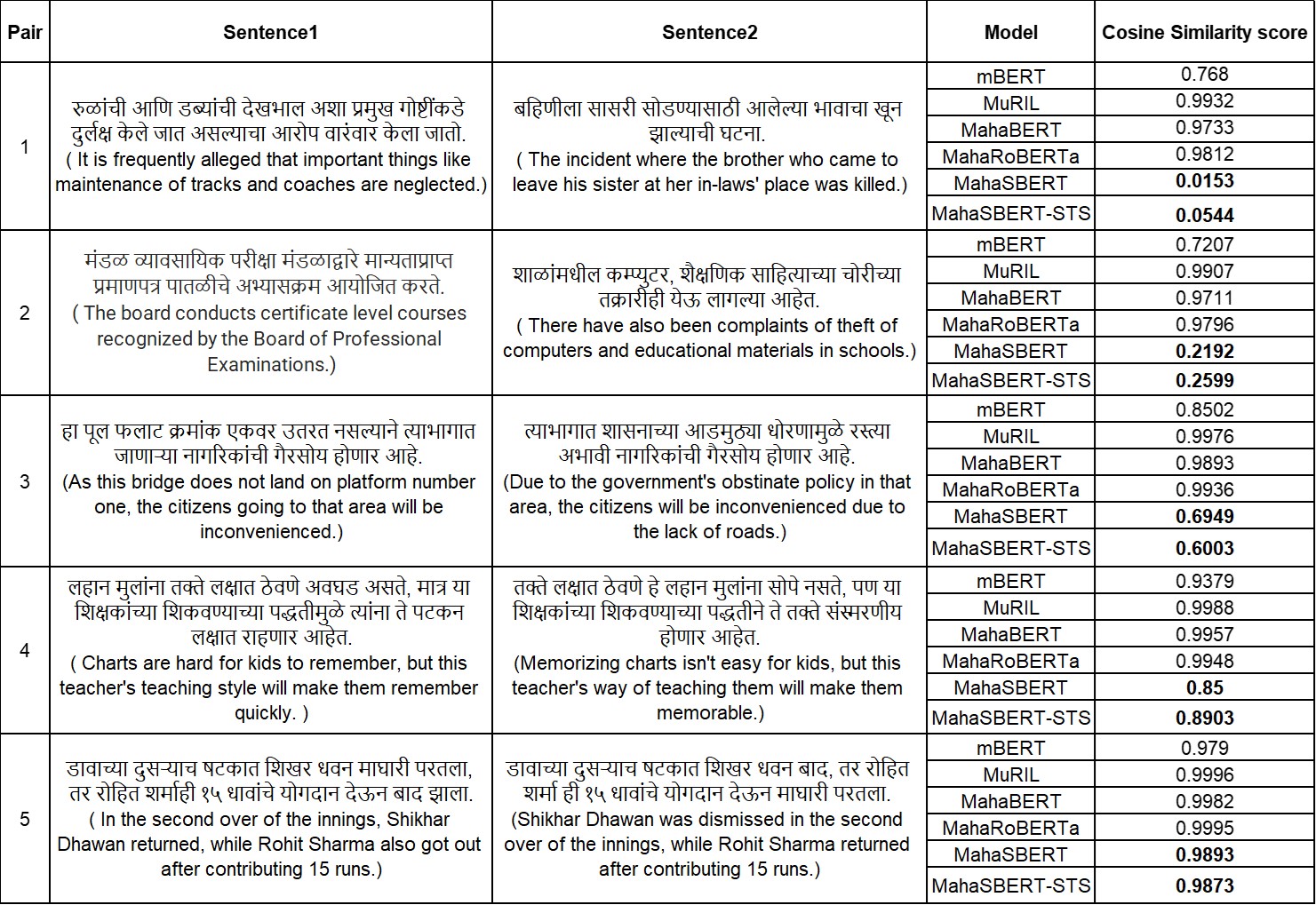}
 \caption{Cosine Similarity scores of various Marathi model embeddings }\label{tab:5}
\end{table}

\begin{table}
\centering
\includegraphics[width=\columnwidth]{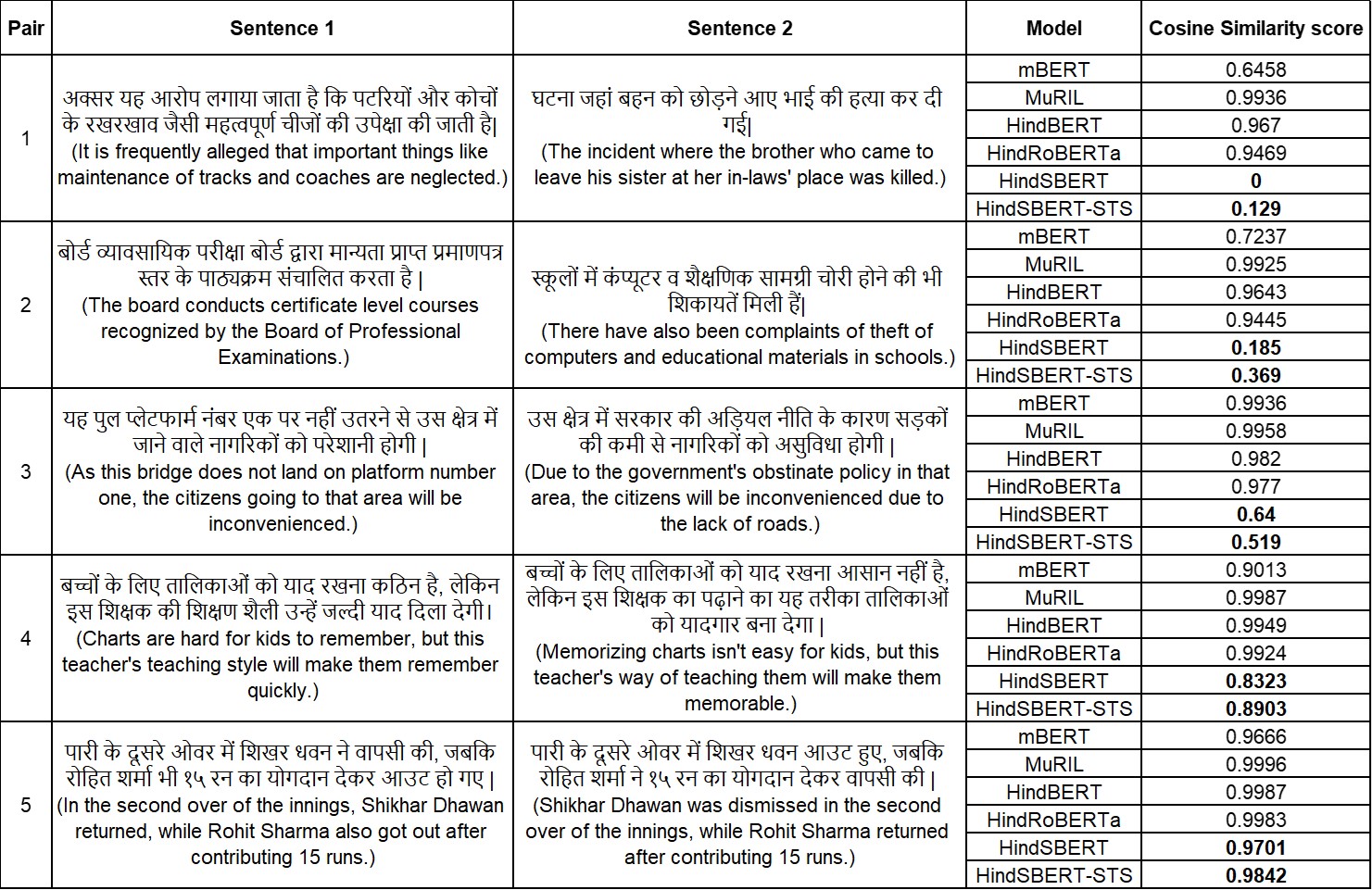}
 \caption{Cosine Similarity scores of various Hindi model embeddings }\label{tab:6}

\end{table}

\subsection{Evaluation results}

Classification accuracies and Embedding Similarity scores have been calculated on three Marathi and three Hindi datasets. The results for the Marathi language are displayed in Table \ref{tab:2 }, while the corresponding results for the Hindi language are displayed in Table \ref{tab:4}. In Table \ref{tab:2 }, the IndicNLP News Articles and iNLTK Headlines datasets are denoted as Articles and Headlines respectively. In Table \ref{tab:4}, the BBC News Articles, IITP- Product reviews and IITP- Movie reviews datasets are termed News-articles, IITP-Products, and IITP-Movies respectively.
\\\\
We find that MahaFT has a slight edge over FB-FT for classification tasks, and performs competitively with the monolingual BERT models.  The monolingual models have either shown comparable performance or outperformed the multilingual versions on all the datasets. This shows the importance of monolingual training for obtaining rich sentence embeddings. However, models having the highest embedding similarity score do not necessarily have the best classification accuracy.
\\\\
We find that Sentence-BERT generates embeddings of a significantly higher quality than FastText and BERT models. Single-step training on STSb (setup 2) produces significantly better performance than single-step training on the NLI dataset (setup 1) as the test data is STSb. For base models, HindBERT and MahaBERT, mean pooling provides a considerable advantage over the CLS pooling strategy. The difference in accuracies between the two pooling strategies is trivial when LaBSE is used as the base model. Thus, the mean pooling strategy is used for the two-step training process (setup 3). Fine-tuning the NLI pre-trained models using STSb has an upper hand over single-step STSb training. Through fine-tuning, a considerable boost in accuracy is achieved for the MuRIL, MahaBERT, and HindBERT base models. The results obtained by mean pooling are presented in Table \ref{tab:3 }.
\\\\
When the three vanilla BERT models of MuRIL, MahaBERT/ HindBERT, and LaBSE are compared, we find that LaBSE gives the best performance, followed by MuRIL and then MahaBERT/ HindBERT. But after applying the two-step training process over these base models, the resultant sentence-BERT models produced by MahaBERT/ HindBERT perform better than those produced by MuRIL and LaBSE.
\\\\
We take a set of 10 sample sentence pairs, chosen randomly from a corpus of news dataset. We use various multilingual and monolingual Hindi and Marathi models to compute the cosine similarity of each sentence pair. The results of this experiment are presented in Table \ref{tab:5} and Table \ref{tab:6}. We observe that the difference in cosine similarity scores provided by the mBERT, MuRIL, MahaBERT, HindBERT, MahaRoBERTa, and HindRoBERTa is insubstantial, thereby making the results non-intuitive. In contrast, the cosine similarity scores obtained from MahaSBERT/ HindSBERT and MahaSBERT-STS/ HindSBERT-STS are intuitive and distinguishable. The difference in the scores of 2 sentence pairs on the opposite ends of the spectrum (exactly similar sentence pair and completely dissimilar sentence pair) is most evident from the embeddings provided by both the Sentence-BERT models. But, the difference is least evident in the similarity scores of embeddings provided by Muril. This points to the need of applying a suitable normalization method on the MuRIL similarity scores to be able to interpret them effectively.
\\\\
Thus, we present the MahaSBERT and HindSBERT: sentence-BERT models trained on the MahaBERT and HindBERT base models respectively. They are trained through the two-step process described in setup 3 above. They give the best performance among all SBERT models of the corresponding language evaluated in this paper. We demonstrate that the Sentence-BERT models made public through this work function competitively with or better than the presently available alternatives for Hindi and Marathi languages.

\section{Conclusion}
In this work, we present a simple approach to training sentence BERT models for low-resource language using the synthetic corpus. We have evaluated these models using a KNN-based classification setup and embedding similarity method on different Hindi and Marathi datasets. Various FastText and pre-trained multilingual and monolingual BERT models have also been evaluated. FastText models are found to perform competitively with monolingual BERT models while the monolingual BERT models outperform multilingual ones. Without any task-specific finetuning, the LaBSE model is found to perform the best for both Hindi and Marathi languages. We highlight the lack of Hindi and Marathi sentence-BERT models in the public domain and hence release MahaSBERT and HindSBERT, the sentence-BERT models created using synthetic datasets. Through a  comparative analysis of their performance, we show that these Sentence-BERT models have an upper hand in the quality of embeddings as compared to all BERT as well as FastText models. They are highly advantageous for the task of semantic sentence similarity. We conclude that the method of two-step training proves to be efficient for developing MahaSBERT-STS and HindSBERT-STS. Finally, we hope that our work facilitates further study and trials in the Hindi and Marathi NLP domains.


\subsubsection{Acknowledgements} 
This work was done under the L3Cube Pune mentorship
program. We would like to express our gratitude towards
our mentors at L3Cube for their continuous support and
encouragement. 

%
%
%
\bibliographystyle{splncs04}
%

\bibliography{main.bib}

\end{document}